\documentclass[11pt,a4paper]{article}
\usepackage[hyperref]{acl2019}
\usepackage{times}
\usepackage{latexsym}

\usepackage{url}

\aclfinalcopy 
\def\aclpaperid{899} 

\usepackage{helvet}  
\usepackage{courier}  
\usepackage{graphicx}  
\usepackage{bm}
\usepackage{amssymb}
\usepackage{amsmath}
\usepackage{array}
\usepackage{hyperref}
\usepackage{float}
\usepackage{multirow}
\usepackage{booktabs}

\newcommand{\E}{{\mathbb{E}}}
\newcommand{\R}{{\mathbb{R}}}

\newcommand{\rbr}[1]{\left(#1\right)}
\newcommand{\sbr}[1]{\left[#1\right]}

\newcolumntype{L}[1]{>{\raggedright\let\newline\\\arraybackslash\hspace{0pt}}m{#1}}
\newcolumntype{C}[1]{>{\centering\let\newline\\\arraybackslash\hspace{0pt}}m{#1}}
\newcolumntype{R}[1]{>{\raggedleft\let\newline\\\arraybackslash\hspace{0pt}}m{#1}}

\title{DIAG-NRE: A Neural Pattern Diagnosis Framework for \\
Distantly Supervised Neural Relation Extraction}

\author{
Shun Zheng$^1$\qquad Xu Han$^2$\qquad Yankai Lin$^2$\qquad Peilin Yu$^3$\qquad Lu Chen$^1$ \\
\textbf{Ling Huang}$^{1,4}$\qquad \textbf{Zhiyuan Liu}$^2$\qquad \textbf{Wei Xu$^1$} \\
$^1$ Institute for Interdisciplinary Information Sciences, Tsinghua University, Beijing, China \\
$^2$ Department of Computer Science and Technology, Tsinghua University, Beijing, China \\
$^3$ Department of Computer Sciences, University of Wisconsin$\textrm{-}$Madison, Madison, USA \\
$^4$ AHI Fintech Inc., Beijing, China \\
\small
\texttt{\{zhengs14,hanxu17,linyk14,lchen17\}@mails.tsinghua.edu.cn; peilin@cs.wisc.edu;} \\
\small
\texttt{linghuang@fintec.ai; \{liuzy,weixu\}@tsinghua.edu.cn;} \\
}

\date{}

\begin{document}

\maketitle

\begin{abstract}
Pattern-based labeling methods have achieved promising results in alleviating the inevitable labeling noises of distantly supervised neural relation extraction. However, these methods require significant expert labor to write relation-specific patterns, which makes them too sophisticated to generalize quickly. To ease the labor-intensive workload of pattern writing and enable the quick generalization to new relation types, we propose a neural pattern diagnosis framework, DIAG-NRE, that can automatically summarize and refine high-quality relational patterns from noise data with human experts in the loop. To demonstrate the effectiveness of DIAG-NRE, we apply it to two real-world datasets and present both significant and interpretable improvements over state-of-the-art methods. Source codes and data can be found at \url{https://github.com/thunlp/DIAG-NRE}.
\end{abstract}

\section{Introduction} 
\label{sec:intro}

Relation extraction aims to extract relational facts from the plain text and can benefit downstream knowledge-driven applications.
A relational fact is defined as a \texttt{relation} between a head entity and a tail entity, e.g., (\textit{Letizia\_Moratti}, \texttt{Birthplace}, \textit{Milan}).
The conventional methods often regard relation extraction as a supervised classification task that predicts the relation type between two detected entities mentioned in a sentence,
including both statistical models~\cite{zelenko2003kernel,zhou2005exploring} and neural models~\cite{zeng2014relation,dossantos2015classify}.

These supervised models require a large number of human-annotated data to train, which are both expensive and time-consuming to collect.
Therefore, \citet{craven1999constructing,mintz2009distant} proposed \textit{distant supervision} (DS) to automatically generate large-scale training data for relation extraction,
by aligning relational facts from a knowledge base (KB) to plain text and assuming that every sentence mentioning two entities can describe their relationships in the KB.
As DS can acquire large-scale data without human annotation,
it has been widely adopted by recent neural relation extraction (NRE) models~\cite{zeng2015distant,lin2016neural}.

\begin{figure}[tb]
\centering
\includegraphics[width=0.95\linewidth]{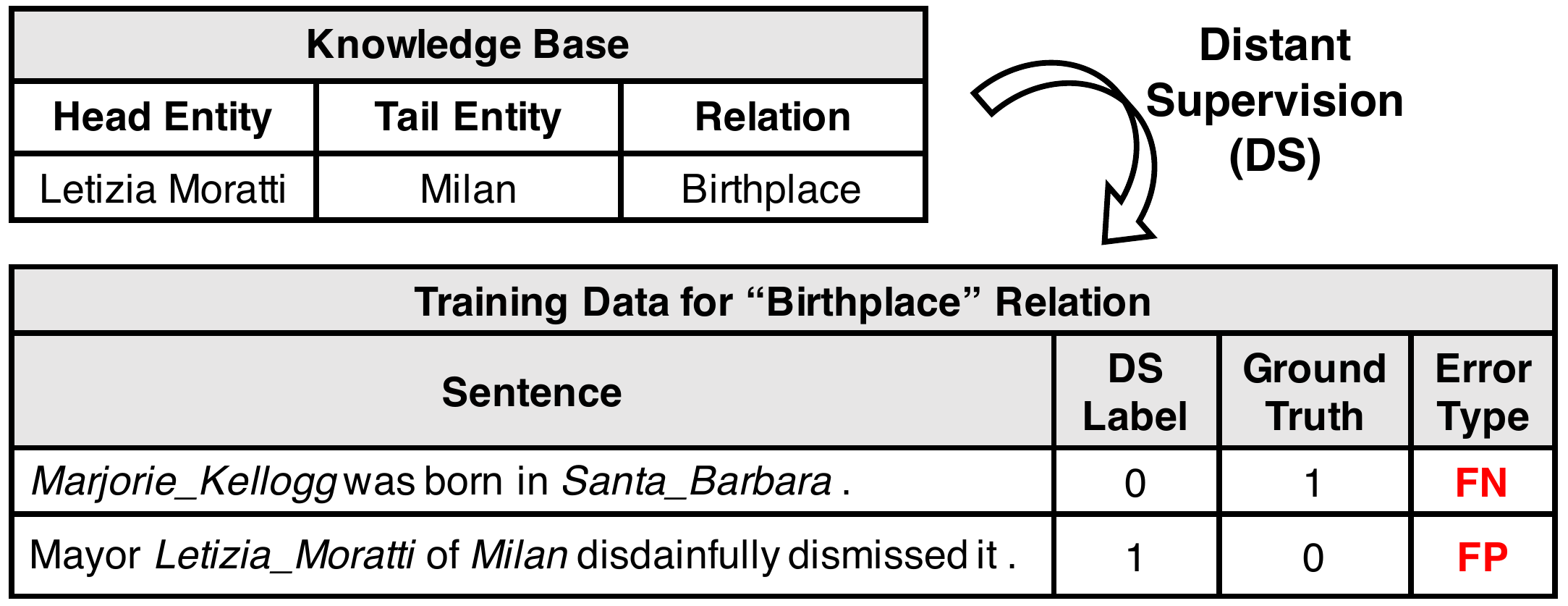}
\caption{Two types of error labels, false negatives (FN) and false positives (FP), caused by DS.}
\label{fig:ds_noise_ex}
\end{figure}

Although DS is both simple and effective in many cases,
it inevitably introduces intolerable labeling noises.
As Figure~\ref{fig:ds_noise_ex} shows, there are two types of error labels, false negatives and false positives.
The reason for false negatives is that a sentence does describe two entities about a target relation, but the fact has not been covered by the KB yet.
While for false positives, it is because not all sentences mentioning entity pairs actually express their relations in the KB.
The noisy-labeling problem can become severe when the KB and text do not match well and as a result heavily weaken the model performance~\cite{riedel2010modeling}.

Recent research has realized that introducing appropriate human efforts is essential for reducing such labeling noises.
For example, \citet{zhang2012big,pershina2014infusion,angeli2014combining,liu2016effective} mixed a small set of crowd-annotated labels with purely DS-generated noise labels.
However, they found that only sufficiently large and high-quality human labels can bring notable improvements, because there are significantly larger number of noise labels.

To enlarge the impact of human efforts, \citet{ratner2016data,liu2017heterogeneous} proposed to incorporate pattern-based labeling, where the key idea was to regard both DS and pattern-based heuristics as the weak supervision sources and develop a weak-label-fusion (WLF) model to produce denoised labels.
However, the major limitation of the WLF paradigm lies in the requirement of human experts to write relation-specific patterns.
Unfortunately, writing good patterns is both a high-skill and labor-intensive task that requires experts to learn detailed pattern-composing instructions, examine adequate examples, tune patterns for different corner cases, etc.
For example, the \texttt{spouse} relation example of~\citet{ratner2016data} uses $11$ functions with over $20$ relation-specific keywords\footnote{\url{https://github.com/HazyResearch/snorkel/tree/master/tutorials/intro}}.
Even worse, when generalizing to a new relation type, we need to repeat the hard manual operations mentioned above again.

\begin{figure}[tb]
\centering
\includegraphics[width=0.95\linewidth]{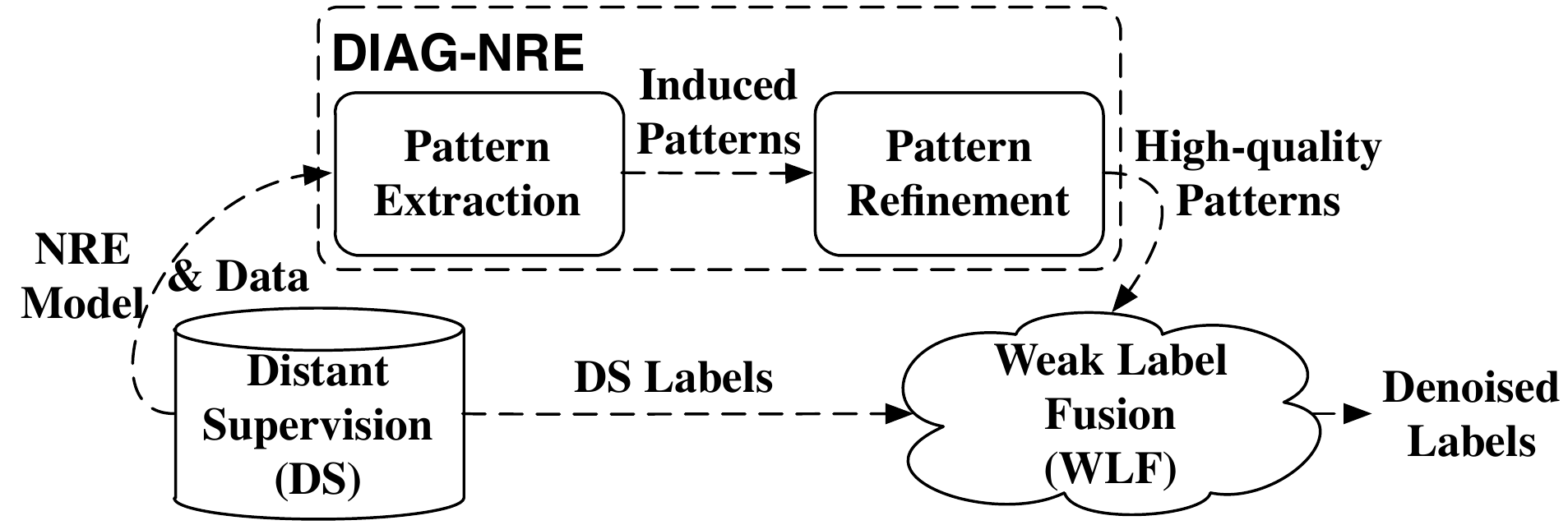}
\caption{An overview of DIAG-NRE.}
\label{fig:overview}
\end{figure}

To ease the pattern-writing work of human experts and enable the quick generalization to new relation types,
we propose a neural pattern diagnosis framework, DIAG-NRE, which establishes a bridge between DS and WLF, for common NRE models.
The general workflow of DIAG-NRE, as Figure~\ref{fig:overview} shows, contains two key stages: 1) \textit{pattern extraction}, extracting potential patterns from NRE models by employing reinforcement learning (RL), and 2) \textit{pattern refinement}, asking human experts to annotate a small set of actively selected examples.
Following these steps, we not only minimize the workload and difficulty of human experts by generating patterns automatically, but also enable the quick generalization by only requiring a small number of human annotations.
After the processing of DIAG-NRE, we obtain high-quality patterns that are either supportive or unsupportive of the target relation with high probabilities and can feed them into the WLF stage to get denoised labels and retrain a better model.
To demonstrate the effectiveness of DIAG-NRE, we conduct extensive experiments on two real-world datasets, where DIAG-NRE not only achieves significant improvements over state-of-the-art methods but also provides insightful diagnostic results for different noise behaviors via refined patterns.

In summary, DIAG-NRE has the following contributions:
\begin{itemize}
    \item easing the pattern-writing work of human experts by generating patterns automatically;
    \item enabling the quick generalization to new relation types by only requiring a small number of human annotations;
    \item presenting both significant and interpretable performance improvements as well as intuitive diagnostic analyses.
\end{itemize}
Particularly, for one relation with severe false negative noises, we improve the F1 score by about $0.4$.
To the best of our knowledge, we are the first to explicitly reveal and address this severe noise problem for that dataset.


\section{Related Work} 
\label{sec:related_work}
To reduce labeling noises of DS, earlier work attempted to design specific model architectures that can better tolerate labeling noises, such as the multi-instance learning paradigm~\cite{riedel2010modeling,hoffmann2011knowledge,surdeanu2012multi,zeng2015distant,lin2016neural,wu2017adversarial}.
These models relax the raw assumption of DS by grouping multiple sentences that mention the same entity pair together as a bag and then assuming that at least one sentence in this bag expresses the relation.
This weaker assumption can alleviate the noisy-labeling problem to some extent,
but this problem still exists at the bag level,
and \citet{feng2018reinforcement} discovered that bag-level models struggled to do sentence-level predictions.

Later work tried to design a dynamic label-adjustment strategy for training~\cite{liu2017soft,luo2017learning}.
Especially, the most recent work~\cite{feng2018reinforcement,qin2018robust} adopted RL to train an agent that interacts with the NRE model to learn how to remove or alter noise labels.
These methods work without human intervention by utilizing the consistency and difference between DS-generated labels and model-predicted ones.
However, such methods can neither discover noise labels that coincide with the model predictions nor explain the reasons for removed or altered labels.
As discussed in the introduction, introducing human efforts is a promising direction to contribute both significant and interpretable improvements, which is also the focus of this paper.

As for the pattern-extraction part,
we note that there are some methods with similar insights but different purposes.
For example, \citet{zhang2018learning} improved the performance of the vanilla LSTM~\cite{hochreiter1997long} by utilizing RL to discover structured representations and \citet{li2016understanding} interpreted the sentiment prediction of neural models by employing RL to find the decision-changing phrases.
However, NRE models are unique because we only care about the semantic inter-entity relation mentioned in the sentence.
To the best of our knowledge, we are the first to extract patterns from NRE models by RL.

We also note that the relational-pattern mining has been extensively studied~\cite{mooney1999relational,carlson2010toward,nakashole2012patty,jiang2017metapad}.
Different from those studies, our pattern-extraction method 1) is simply based on RL, 2) does not rely on any lexical or syntactic annotation, and 3) can be aware of the pattern importance via the prediction of NRE models.
Besides,~\citet{takamatsu2012reducing} inferred negative syntactic patterns via the example-pattern-relation co-occurrence and removed the false-positive labels accordingly.
In contrast, built upon modern neural models, our method not only reduces negative patterns to alleviate false positives but also reinforces positive patterns to address false negatives at the same time.


\section{Methodology} 
\label{sec:methodology}

Provided with DS-generated data and NRE models trained on them, DIAG-NRE can generate high-quality patterns for the WLF stage to produce denoised labels.
As Figure~\ref{fig:overview} shows, DIAG-NRE contains two key stages in general: pattern extraction (Section~\ref{sec:deep_pat_ext}) and pattern refinement (Section \ref{sec:pat_ref}).
Moreover, we briefly introduce the WLF paradigm in Section~\ref{sec:wlf} for completeness.
Next, we start with reviewing the common input-output schema of modern NRE models.

\subsection{NRE Models}
\label{sec:nre}
Given an instance $s$ with $T$ tokens\footnote{In this paper, we refer to a sentence together with an entity pair as an instance and omit the instance index for brevity.},
a common input representation of NRE models is $\bm{x} = [\bm{x}_1, \bm{x}_2,\cdots, \bm{x}_T]$,
where $\bm{x}_i \in \R^{d^x}$ denotes the embedding of token $i$ and $d^x$ is the token embedding size.
Particularly, $\bm{x}_i$ is the concatenation of the word embedding, $\bm{w}_i \in \R^{d^x}$, and position embedding, $\bm{p}_i \in \R^{d^p}$, as $[\bm{w}_i; \bm{p}_i]$, to be aware of both semantics and entity positions, where $d^x = d^w + d^p$.
Given the relation type $r$,
NRE models perform different types of tensor manipulations on $\bm{x}$
and obtain the predicting probability of $r$ given the instance $s$ as $P_\phi(r|\bm{x})$,
where $\phi$ denotes model parameters except for the input embedding tables.

\subsection{Pattern Extraction}
\label{sec:deep_pat_ext}
\begin{figure}[tb]
\centering
\includegraphics[width=0.95\linewidth]{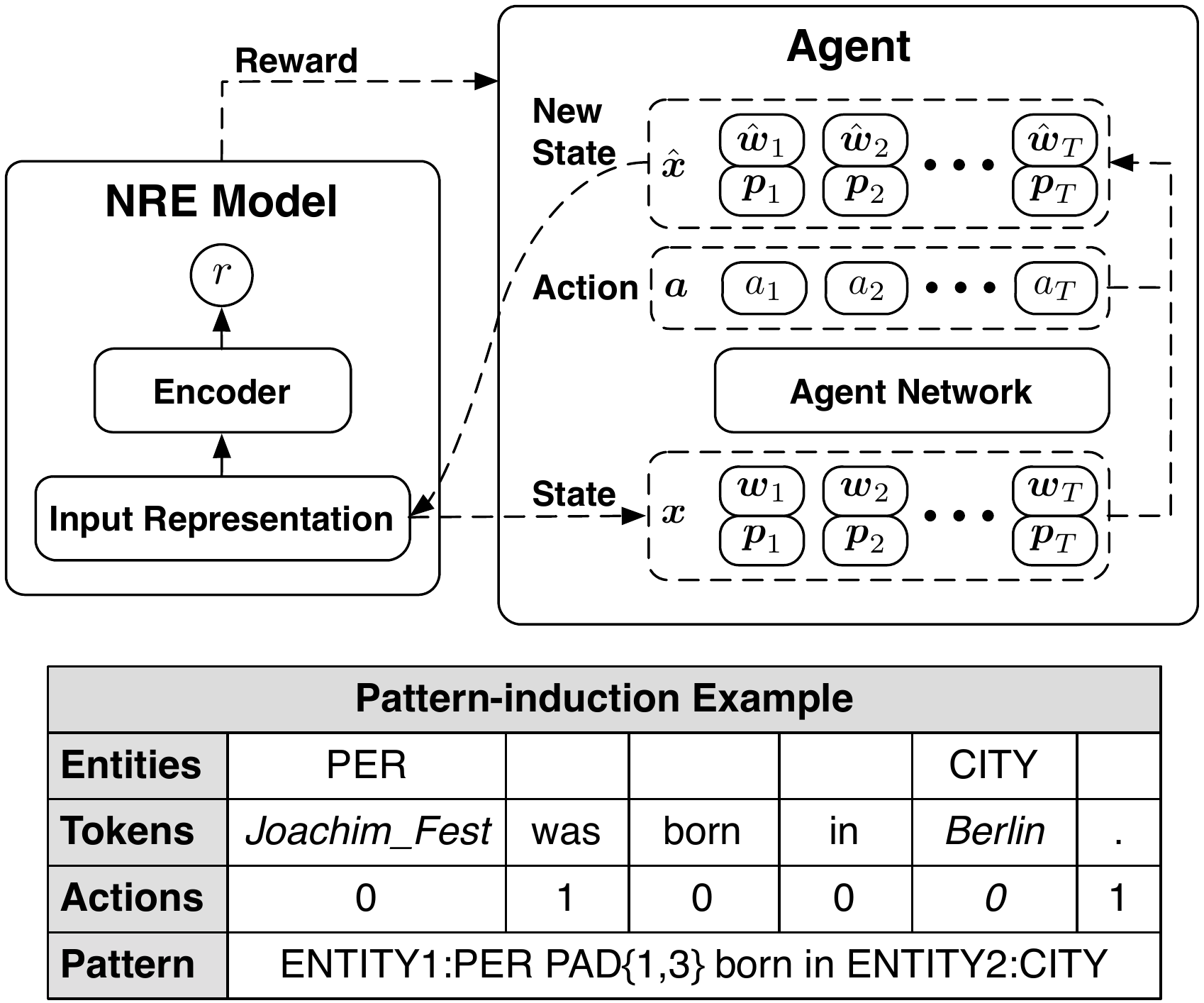}
\caption{The RL-based pattern-extraction workflow and a typical pattern-induction example, where we induce a pattern for the \texttt{Birthplace} relation via a series of actions ($0$: \textit{retaining}, $1$: \textit{erasing}).}
\label{fig:pat_ext}
\end{figure}

In this stage, we build a pattern-extraction agent to distill relation-specific patterns from NRE models with the aforementioned input-output schema.
The basic idea is to erase irrelevant tokens and preserve the raw target prediction simultaneously, which can be modeled as a token-erasing decision process and optimized by RL.
Figure~\ref{fig:pat_ext} shows this RL-based workflow in a general way together with an intuitive pattern-induction example.
Next, we elaborate details of this workflow.

\paragraph{Action.}
The agent takes an action $a_i$, \textit{retaining} (0) or \textit{erasing} (1), for each token of the instance $s$ and transforms the input representation from $\bm{x}$ into $\bm{\hat{x}}$.
During this process, the column $i$ of $\bm{x}$, $\bm{x}_i = [\bm{w}_i; \bm{p}_i]$,
corresponding to the token $i$ of raw instance $s$,
is transformed into $\hat{\bm{x}}_i = [\hat{\bm{w}}_i; \bm{p}_i]$,
where the position vectors are left untouched and the new word vector $\hat{\bm{w}}_i$ is adjusted based on the action taken by the agent.
For the \textit{retaining} action, we retain the raw word vector as $\hat{\bm{w}}_i = \bm{w}_i$.
While for \textit{erasing}, we set $\hat{\bm{w}}_i$ to be all zeros to remove the semantic meaning.
After taking a sequence of actions, $\bm{a} = [a_1; a_2; \cdots; a_T]$,
we get the transformed representation $\hat{\bm{x}}$ with $\hat{T}$ tokens retained.

\paragraph{Reward.}
Our purpose is to find the most simplified sequence that preserves the raw prediction confidence.
Therefore, given the raw input representation $x$ and the corresponding action vector $\bm{a}$,
we define the reward as follows:
\begin{align*}
    R(\bm{a}|\bm{x}) = \underbrace{\log \left( \frac{P_\phi(r|\bm{\hat{x}})}{P_\phi(r|\bm{x})} \right) }_{\text{Prediction Confidence}} + \eta \cdot 
    \underbrace{(1 - \Hat{T}/T)}_{\text{Sparsity}},
\end{align*}
where the total reward is composed of two parts: one is the log-likelihood term to pursue the high prediction confidence and the other is the sparse ratio term to induce sparsity in terms of retained tokens.
We balance these two parts through a hyper-parameter $\eta$.

\paragraph{State.}
To be general, the state provided to the agent should be independent of NRE architectures.
Moreover, the state needs to incorporate complete information of the current instance.
Therefore, in our design, the agent directly employs the input representation $\bm{x}$ as the state.

\paragraph{Agent.}
We employ policy-based RL to train a neural-network-based agent that can predict a sequence of actions for an instance to maximize the reward.
Our agent network directly estimates $\pi_\Theta(\bm{a}|\bm{x}) = \prod_{i=1}^T \pi_\Theta(a_i|\bm{x})$ in a non-autoregressive manner by calculating $\pi_\Theta(a_i|\bm{x})$ in parallel, where $\Theta$ denotes the parameters of the agent network.
To enrich the contextual information when deciding the action for each token, we employ the forward and backward LSTM networks to encode $\bm{x}$ into $\bm{h}$ as
\begin{align*}
    \overrightarrow{\bm{h}}
    &= [\overrightarrow{h}_1, \overrightarrow{h}_2, \cdots, \overrightarrow{h}_T ]
    = \text{Forward-LSTM}(\bm{x}), \\
    \overleftarrow{\bm{h}}
    &= [\overleftarrow{h}_1, \overleftarrow{h}_2, \cdots, \overleftarrow{h}_T ]
    = \text{Backward-LSTM}(\bm{x}), \\
    \bm{h}
    &= [h_1, h_2, \cdots, h_T ] 
    = \text{Concatenate}(\overrightarrow{\bm{h}}, \overleftarrow{\bm{h}}),
\end{align*}
where $\overrightarrow{h}_i \in \R^{d^h}$, $\overleftarrow{h}_i \in \R^{d^h}$, $h_i = [\overrightarrow{h}_i; \overleftarrow{h}_i] \in \R^{2\times d^h}$,
and $d^h$ denotes the size of LSTM's hidden state.
Then, we employ an attention-based strategy~\cite{bahdanau2015neural} to aggregate the contextual information as $\bm{c} = [c_1, c_2, \cdots, c_T]$.
For each token $i$, we compute the context vector $c_i \in \R^{2d^h}$ as follows:
\begin{align*}
    c_i = \sum_{j=1}^T \alpha_j^i h_j,
\end{align*}
where each scalar weight $\alpha_j^i$ is calculated by $e_j^i / (\sum_{k=1}^T e_k^i)$.
Here $e_j^i$ is computed by a small network as $ e_j^i = v_\alpha^\top \tanh(W_x x_i + W_h h_j)$,
where $W_x \in \R^{2d_h\times d_x}$, $W_h \in \R^{2d_h\times 2d_h}$ and $v_\alpha \in \R^{2d_h}$ are network parameters.
Next, we compute the final representation to infer actions as $\bm{z} = [z_1, z_2, \cdots, z_T]$,
where for each token $i$, $z_i = [x_i; c_i] \in \R^{d^x+2d^h}$ incorporates semantic, positional and contextual information.
Finally, we estimate the probability of taking action $a_i$ for token $i$ as
\begin{align*}
    \pi_\Theta(a_i|\bm{x}) &= o_i^{a_i} \cdot (1-o_i)^{(1-a_i)},
\end{align*}
where $o_i = \text{sigmoid}(W_o^\top z_i + b_o)$, $W_o \in \R^{d^x + 2d^h}$ and $b_o \in \R^1$ are network parameters.

\paragraph{Optimization.}
We employ the REINFORCE algorithm~\cite{williams1992simple} and policy gradient methods~\cite{sutton2000policy} to optimize parameters of the agent network,
where the key step is to rewrite the gradient formulation and then apply the back-propagation algorithm~\cite{rumelhart1986learning} to update network parameters.
Specifically, we define our objective as:
\begin{align*}
    L(\Theta) = \E_s\sbr{\E_{\pi_\Theta(\bm{a}|\bm{x})} R(\bm{a}|\bm{x})},
\end{align*}
where $\bm{x}$ denotes the input representation of the instance $s$.
By taking the derivative of $J(\Theta)$ with respect to $\Theta$, we can obtain the gradient $\nabla_\Theta L(\Theta)$ as $\E_s[ \E_{\pi_\Theta(\bm{a}|\bm{x})}[ R(\bm{a}|\bm{x}) \nabla_\Theta \log \pi_\Theta(\bm{a}|\bm{x}) ]]$.
Besides, we utilize the $\epsilon$-greedy trick to balance exploration and exploitation.

\paragraph{Pattern Induction.}
Given instances and corresponding agent actions,
we take the following steps to induce compact patterns.
First, to be general, we substitute raw entity pairs with corresponding entity types.
Then, we evaluate the agent to obtain retained tokens with the relative distance preserved.
To enable the generalized position indication,
we divide the relative distance between two adjacent retained tokens into four categories:
zero (no tokens between them), short (1-3 tokens),
medium (4-9 tokens) and long (10 or more tokens) distance.
For instance, Figure~\ref{fig:pat_ext} shows a typical pattern-induction example.
Patterns with such formats can incorporate multiple kinds of crucial information, such as entity types, key tokens and the relative distance among them.

\subsection{Pattern Refinement}
\label{sec:pat_ref}

\begin{figure}[tb]
\centering
\includegraphics[width=0.95\linewidth]{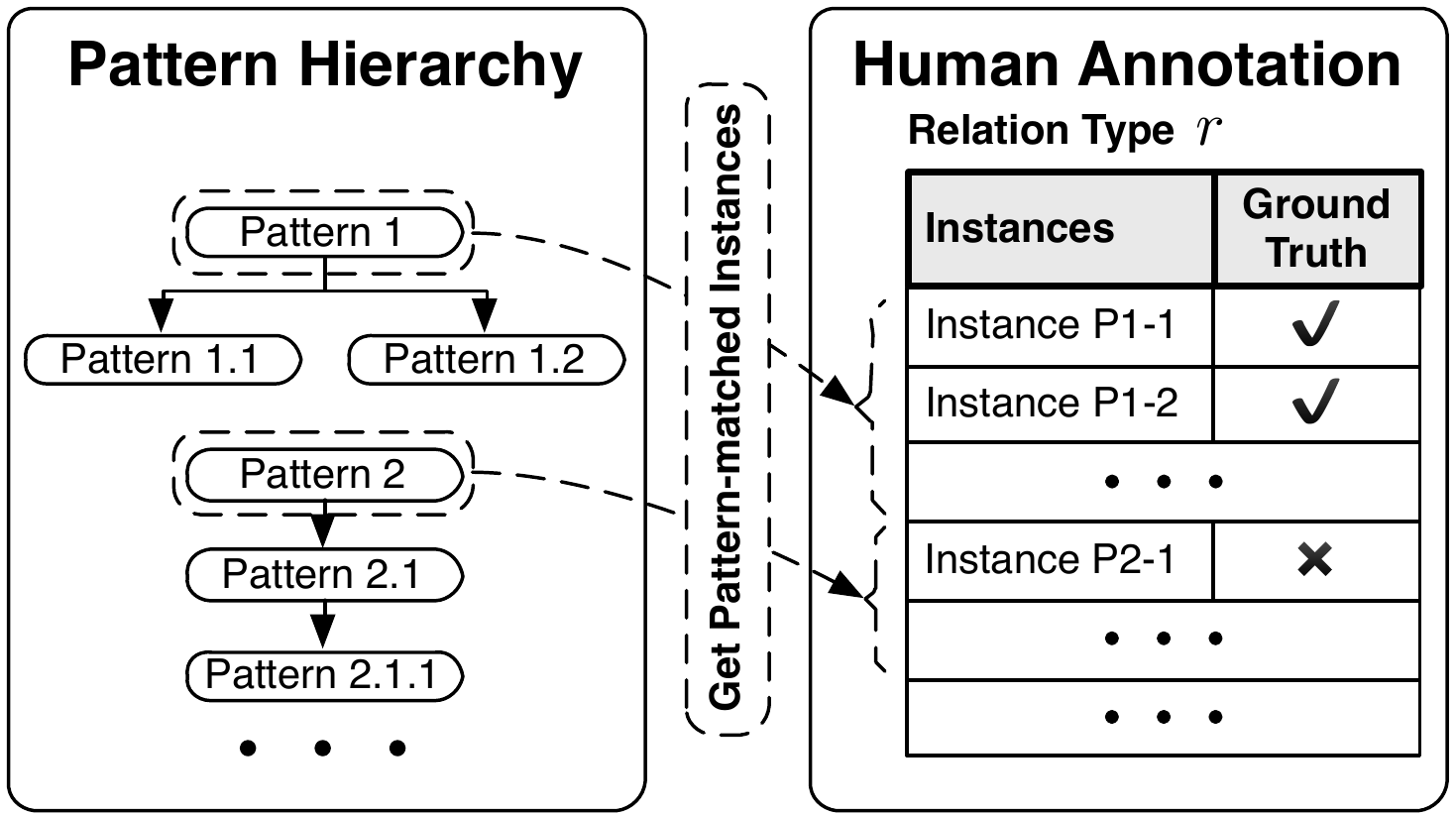}
\caption{The human-in-the-loop pattern refinement.}
\label{fig:pat_ref}
\end{figure}

The above pattern-extraction stage operates at the instance level by producing a pattern for each evaluated instance.
However, after aggregating available patterns at the dataset level,
there inevitably exist redundant ones.
Therefore, we design a pattern hierarchy to merge redundant patterns.
Afterward, we can introduce human experts into the workflow by asking them to annotate a small number of actively selected examples.
Figure~\ref{fig:pat_ref} shows the general workflow of this stage.

\paragraph{Pattern Hierarchy.}
To identify redundant patterns, 
we group multiple instances with the same pattern and build a pattern hierarchy by the matching statistics.
In this hierarchy, the parent pattern should cover all instances matched by the child pattern.
As the parent pattern already has sufficient relation-supporting signals, we can omit child patterns for human annotation.
Moreover, the number of instances from which the pattern can be induced is closely related to the pattern representativeness.
Therefore, we follow the decreasing order of this number to select top $n_r$ most representative patterns for human annotation.

\paragraph{Human Annotation.}
To quantitatively evaluate the pattern quality, we adopt an approximate method by randomly selecting $n_a$ pattern-matched instances and annotating them manually.
Thus, for each relation type, we end up with $n_r*n_a$ human-annotated instances.
We assign patterns with the accuracy higher than $p_h$ and lower than $p_l$ into the positive pattern set and the negative pattern set, respectively, to serve the WLF stage.
In practice, users can tune these hyper-parameters ($n_r$, $n_a$, $p_h$ and $p_l$) accordingly for different applications, such as increasing $p_h$ to prefer precision.
While in this paper, to show the wide applicability and robustness of DIAG-NRE, we demonstrate that a single configuration can handle all 14 relation types.

\subsection{Weak Label Fusion}
\label{sec:wlf}
The WLF model aims to fuse weak labels from multiple labeling sources, including both DS and patterns, to produce denoised labels.
In this paper, we adopt \textit{data programming} (DP)~\cite{ratner2016data} at our WLF model.
The input unit of DP is called \textit{labeling function} (LF), which takes one instance and emits a label (+1: \textit{positive}, -1: \textit{negative} or 0: \textit{unknown}).
In our case, the LF of DS generates +1 or -1, LFs of positive patterns generate +1 or 0,
and LFs of negative patterns generate -1 or 0.
We estimate parameters of DP on the small set of human-annotated labels with a closed-form solution (see the appendix for detailed formulations).
With the help of DP, we get denoised labels to retrain a better model.
Note that designing better generic WLF models is still a hot research topic~\cite{varma2016socratic,bach2017learning,liu2017heterogeneous} but outside the scope of this work, which is automatically generating patters to ease human's work.


\section{Experiments} 
\label{sec:experiments}
In this section, we present experimental results and comprehensive analyses to demonstrate the effectiveness of DIAG-NRE.

\subsection{Experimental Setup}

\paragraph{Evaluation.}
To clearly show the different noise behaviours for various relation types, we treat each relation prediction task as a single binary classification problem, that is predicting the existing or not of that relation for a given instance.
Different from previous studies, we report relation-specific metrics (Precision, Recall and F1 scores, all in the percentage format) and macro-averaged ones at the dataset level, because the distribution of relation types is extremely imbalanced and the micro-averaged evaluation inevitably overlooks noisy-labeling issues of many relation types.
Moreover, we only utilize human-annotated test data to evaluate models trained on noise labels, as~\citet{ratner2016data,liu2016effective} did.
The reason is that the severe labeling noises of many relation types heavily weaken the reliability of the DS-based \textit{heldout evaluation}~\cite{mintz2009distant}, which cannot judge the performance accurately.

\begin{table}[tb]\small
\centering
\begin{tabular}{c|c|l|r|r}
\toprule
 & \textbf{TID} & \textbf{Relation Abbreviation} & \textbf{Train} & \textbf{Test} \\
\midrule

\multirow{10}{*}{\rotatebox{270}{NYT}}
& $R_0$ & \texttt{Bus.\slash Company}      & $5.3$k  & $186$ \\ 
& $R_1$ & \texttt{Loc.\slash Admin.\_Div.} & $4.9$k  & $180$ \\
& $R_2$ & \texttt{Loc.\slash Capital}      & $5.3$k  & $20 $ \\
& $R_3$ & \texttt{Loc.\slash Contains}     & $44.6$k & $263$ \\  
& $R_4$ & \texttt{Loc.\slash Country}      & $4.9$k  & $89 $ \\  
& $R_5$ & \texttt{Loc.\slash Neighbor.}    & $5.6$k  & $55 $ \\ 
& $R_6$ & \texttt{Peo.\slash National.}    & $7.5$k  & $84 $ \\
& $R_7$ & \texttt{Peo.\slash Place\_Lived} & $6.7$k  & $230$ \\ 
& $R_8$ & \texttt{Peo.\slash Birthplace}   & $3.1$k  & $16 $ \\
& $R_9$ & \texttt{Peo.\slash Deathplace}   & $1.9$k  & $19 $ \\  
\midrule
\multirow{4}{*}{\rotatebox{270}{UW}}
& $R^u_6$ & \texttt{Peo.\slash National.} & $107$k  & $1.8$k \\
& $R^u_7$ & \texttt{Peo.\slash Place\_Lived} & $20.9$k & $3.8$k \\ 
& $R^u_8$ & \texttt{Peo.\slash Birthplace} & $15.3$k & $458$ \\
& $R^u_9$ & \texttt{Peo.\slash Deathplace} & $5.7$k  & $1.3$k \\  

\bottomrule
\end{tabular}
\caption{The total $14$ relation prediction tasks with corresponding task IDs (TIDs), relation abbreviations and the number of positive labels in the train and test sets.
The train set, generated by DS, contains $452,223$ and $395,738$ instances for NYT and UW, respectively.
The test set, annotated by the human, contains $1,027$ and $15,622$ instances for NYT and UW, respectively.}
\label{table:relations}
\end{table}

\paragraph{Data \& Tasks.}
We select top ten relation types with enough coverage (over $1,000$ instances) from the NYT dataset~\cite{riedel2010modeling}\footnote{\url{http://iesl.cs.umass.edu/riedel/ecml/}} and all four relation types from the UW dataset~\cite{liu2016effective}\footnote{\url{https://www.cs.washington.edu/ai/gated_instructions/naacl_data.zip}}.
Originally, the NYT dataset contains a train set and a test set both generated by DS with $522,611$ and $172,448$ instances, respectively;
the UW dataset contains a train set generated by DS, a crowd-annotated set and a minimal human-annotated test set with $676,882$, $18,128$ and $164$ instances, respectively.
To enable the reliable evaluation based on human annotations,
for the NYT dataset, we randomly select up to $100$ instances per relation (including the special unknown relation \texttt{NA}) from the test set and manually annotate them;
while for the UW dataset,  we directly utilize the crowd-annotated set (disjoint from the train set) with the broad coverage and very high quality as the ground truth.
Table~\ref{table:relations} summaries detailed statistics of these $14$ tasks.

\paragraph{Hyper-parameters.}
We implement DIAG-NRE based on Pytorch\footnote{\url{https://pytorch.org/}} and directly utilize its default initialization for neural networks.
For the NRE model, we adopt a simple yet effective LSTM-based architecture described in~\citet{zhou2016attention} and adopt widely-used hyper-parameters (see the appendix for details).
As for DIAG-NRE, we use the following configuration for all $14$ tasks.
For the agent network, the LSTM hidden size is $200$,
the optimizer is Adam with a learning rate of $0.001$,
the batch size is $5$, and the training epoch is $10$.
At the pattern-extraction stage, we use $\epsilon = 0.1$ and alter $\eta$ in $\{0.05, 0.1, 0.5, 1.0, 1.5\}$ to train multiple agents that tend to squeeze patterns with different granularities and combine outputs of all agents to serve the pattern-refinement stage.
To speed up the agent training, we use filtered instances by taking the top $10,000$ ones with the highest prediction probabilities.
At the pattern-refinement stage, hyper-parameters include $n_r=20$, $n_a=10$, $p_h=0.8$ and $p_l=0.1$.
Thus, for each task, we get $200$ human-annotated instances (about $0.05\%$ of the entire train set) and at most $20$ patterns for the WLF stage.


\begin{table*}[tb]\small
\centering
\begin{tabular}{c| ccc | ccc | ccc | cccrr}

\toprule
\multirow{2}{*}{\textbf{TID}}
& \multicolumn{3}{c|}{\textbf{Distant Supervision}}
& \multicolumn{3}{c|}{\textbf{Gold Label Mix}}
& \multicolumn{3}{c|}{\textbf{RLRE}}
& \multicolumn{5}{c}{\textbf{DIAG-NRE}}
\\
& \textbf{P.} & \textbf{R.} & \textbf{F1}
& \textbf{P.} & \textbf{R.} & \textbf{F1}
& \textbf{P.} & \textbf{R.} & \textbf{F1}
& \textbf{P.} & \textbf{R.} & \textbf{F1} & \textbf{Inc-DS} & \textbf{Inc-Best}
\\
\midrule
$R_{0}$ & 95.1 & 41.5 & 57.8 & 95.7 & 40.8 & 57.2 & 97.7 & 32.4 & 48.6 & 95.7 & 42.8 & \textbf{59.1} & \textbf{+1.4} & \textbf{+1.4} \\ 

$R_{1}$ & 91.9 & 9.1 & 16.4 & 90.2 & 11.7 & 20.2 & 92.6 & 4.2 & 8.0 & 94.5 & 44.8 & \textbf{60.7} & \textbf{+44.3} & \textbf{+40.4} \\ 

$R_{2}$ & 37.0 & 83.0 & 50.8 & 40.0 & 85.0 & 54.0 & 64.8 & 68.0 & \textbf{66.1} & 42.4 & 85.0 & 56.0 & \textbf{+5.2} & -10.1 \\ 

$R_{3}$ & 87.5 & 79.2 & 83.2 & 87.1 & 80.2 & \textbf{83.5} & 87.5 & 79.2 & 83.2 & 87.0 & 79.8 & 83.2 & +0.0 & -0.3 \\ 

$R_{4}$ & 95.3 & 50.1 & 64.7 & 94.1 & 49.0 & 63.9 & 98.2 & 47.9 & 64.0 & 94.5 & 57.5 & \textbf{71.5} & \textbf{+6.7} & \textbf{+6.7} \\ 

$R_{5}$ & 82.7 & 29.1 & 42.9 & 84.7 & 29.5 & 43.6 & 82.7 & 29.1 & 42.9 & 84.5 & 37.5 & \textbf{51.8} & \textbf{+8.9} & \textbf{+8.3} \\ 

$R_{6}$ & 82.0 & 83.8 & \textbf{82.8} & 81.6 & 84.0 & 82.7 & 82.0 & 83.8 & 82.8 & 81.5 & 83.3 & 82.3 & -0.5 & -0.5 \\ 

$R_{7}$ & 82.3 & 22.3 & 35.1 & 82.0 & 22.6 & 35.4 & 83.5 & 21.8 & 34.5 & 82.0 & 25.6 & \textbf{39.0} & \textbf{+3.8} & \textbf{+3.6} \\ 

$R_{8}$ & 66.2 & 32.5 & 39.8 & 70.5 & 47.5 & 55.8 & 66.2 & 32.5 & 39.8 & 73.4 & 61.3 & \textbf{65.5} & \textbf{+25.7} & \textbf{+9.7} \\ 

$R_{9}$ & 85.4 & 73.7 & 77.9 & 85.9 & 80.0 & 81.5 & 85.4 & 73.7 & 77.9 & 89.0 & 87.4 & \textbf{87.1} & \textbf{+9.2} & \textbf{+5.6} \\ 

Avg. & 80.5 & 50.4 & 55.1 & 81.2 & 53.0 & 57.8 & 84.1 & 47.3 & 54.8 & 82.5 & 60.5 & \textbf{65.6} & \textbf{+10.5} & \textbf{+6.5} \\

\midrule

$R^u_{6}$ & 35.9 & 75.7 & 48.7 & 35.8 & 75.0 & 48.5 & 36.0 & 75.3 & \textbf{48.7} & 36.2 & 74.5 & 48.7 & +0.0 & -0.0 \\ 

$R^u_{7}$ & 57.8 & 18.5 & 28.0 & 59.3 & 19.1 & 28.8 & 57.8 & 18.5 & 28.0 & 56.3 & 23.5 & \textbf{33.1} & \textbf{+5.1} & \textbf{+4.3} \\ 

$R^u_{8}$ & 37.3 & 64.0 & 46.9 & 40.0 & 64.9 & 49.1 & 37.3 & 64.0 & 46.9 & 48.1 & 71.9 & \textbf{57.5} & \textbf{+10.6} & \textbf{+8.3} \\ 

$R^u_{9}$ & 77.1 & 71.3 & 74.0 & 77.5 & 70.3 & 73.5 & 77.1 & 71.3 & 74.0 & 80.7 & 71.1 & \textbf{75.4} & \textbf{+1.5} & \textbf{+1.5} \\ 

Avg. & 52.0 & 57.4 & 49.4 & 53.1 & 57.3 & 50.0 & 52.0 & 57.3 & 49.4 & 55.3 & 60.2 & \textbf{53.7} & \textbf{+4.3} & \textbf{+3.5} \\
 
\midrule

\end{tabular}
\caption{Overall results for $14$ tasks, where we present relation-specific scores, the macro-averaged ones (Avg.), the F1 improvement of DIAG-NRE over the vanilla DS (Inc-DS) and the best baseline (Inc-Best), and we highlight the best F1 for each task and the significant improvements.}
\label{table:experiment}
\end{table*}

\begin{table}[tb]\small
\centering
\begin{tabular}{c r r r r r}
\toprule
\textbf{TID} & \textbf{Prec.} & \textbf{Recall} & \textbf{Acc.} & \textbf{\#Pos.} & \textbf{\#Neg.} \\
\midrule
$R_0$   &100.0        &81.8         &82.0         &20 &0  \\ 
$R_1$   &93.9         &\underline{33.5} & \underline{36.2} &18 &0  \\
$R_2$   &75.7         &88.0         &76.5         &9  &5  \\
$R_3$   &100.0        &91.4         &92.0         &20 &0  \\ 
$R_4$   &93.3         &72.4         &80.9         &10 &2  \\
$R_5$   &93.8         &77.3         &86.5         &15 &0  \\ 
$R_6$   &88.3         &76.9         &75.1         &14 &0  \\
$R_7$   &91.9         &64.6         &64.0         &20 &0  \\ 
$R_8$   &\underline{29.3} & \underline{30.4} & 60.0         &4  &10 \\
$R_9$   &66.7         & \underline{38.1} &74.4         &6  &11 \\
$R^u_6$ &81.8         &90.7         &81.0         &7  &0  \\
$R^u_7$ &93.5         &70.7         &68.3         &17 &1  \\
$R^u_8$ & \underline{35.0} &70.0         &60.0         &4  &15 \\
$R^u_9$ &87.5         &59.2         &67.7         &12 &5  \\ 
\bottomrule
\end{tabular}
\caption{Total diagnostic results, where columns contain the precision, recall and accuracy of DS-generated labels evaluated on 200 human-annotated labels as well as the number of positive and negative patterns preserved after the pattern-refinement stage, and we underline some cases in which DS performs poorly.}
\label{table:diagnosis}
\end{table}

\begin{table*}[tb]\small
\centering
\begin{tabular}{c L{10cm} c c c}
\toprule
\textbf{TID} & \textbf{Patterns \& Matched Examples} & \textbf{DS} & \textbf{RLRE} & \textbf{DIAG-NRE} \\
\midrule

\multirow{3}{*}{$R_1$} &
\multicolumn{4}{l}{
\textbf{Pos. Pattern:} \texttt{in ENTITY2:CITY PAD\{1,3\} ENTITY1:COUNTRY}
\quad
(\textbf{DS Label:} 382 / 2072)
} \\
& \textbf{Example:}
He will , however , perform this month in \textit{Rotterdam} , the \textit{Netherlands} , and Prague . & 0 & \texttt{None} & 0.81 \\

\midrule

\multirow{4}{*}{$R_8$} &
\multicolumn{4}{l}{
\textbf{Pos. Pattern}: \texttt{ENTITY1:PER PAD\{1,3\} born PAD\{1,3\} ENTITY2:CITY} 
\quad
(\textbf{DS Label:} 44 / 82)
} \\
& \textbf{Example:} \textit{Marjorie\_Kellogg} was born in \textit{Santa\_Barbara} .  & 0 & 0 & 1.0 \\
& \multicolumn{4}{l}{
\textbf{Neg. Pattern:} \texttt{mayor ENTITY1:PER PAD\{1,3\} ENTITY2:CITY} 
\quad
(\textbf{DS Label:} 21 / 62)
} \\
& \textbf{Example:} Mayor \textit{Letizia\_Moratti} of \textit{Milan} disdainfully dismissed it . & 1 & 1 & 0.0 \\

\midrule

\multirow{4}{*}{$R^u_9$} &
\multicolumn{4}{l}{
\textbf{Pos. Pattern:} \texttt{ENTITY1:PER died PAD\{4,9\} ENTITY2:CITY} 
\quad
(\textbf{DS Label:} 66 / 108)
} \\
& \textbf{Example:} \textit{Dahm} died Thursday at an assisted living center in \textit{Huntsville}
...
& 0 & 0 & 1.0 \\
& \multicolumn{4}{l}{
\textbf{Neg. Pattern:} \texttt{ENTITY1:PER PAD\{4,9\} rally PAD\{1,3\} ENTITY2:CITY} 
\quad
(\textbf{DS Label:} 40 / 87)
} \\
& \textbf{Example:} \textit{Bhutto} vowed to hold a rally in \textit{Rawalpindi} on Friday
...
& 1 & 1 & 0.0 \\

\bottomrule
\end{tabular}
\caption{Positive (Pos.), negative (Neg.) patterns and associated examples with labels produced by different methods.
For each pattern, we present ``DS Label'' as the number of DS-generated positive labels over the number of pattern-matched instances.
For RLRE, \texttt{None} means the instance is removed. 
For DIAG-NRE, we present the soft label produced by the WLF model.}
\label{table:case}
\end{table*}

\subsection{Performance Comparisons}

Based on the above hyper-parameters,
DIAG-NRE together with the WLF model can produce denoised labels to retrain a better NRE model.
Next, we present the overall performance comparisons of NRE models trained with different labels.

\paragraph{Baselines.}
We adopt the following baselines:
1) \textit{Distant Supervision}, the vanilla DS described in~\citet{mintz2009distant},
2) \textit{Gold Label Mix}~\cite{liu2016effective}, mixing human-annotated high-quality labels with DS-generated noise labels,
and 3) \textit{RLRE}~\cite{feng2018reinforcement}, building an instance-selection agent to select correct-labeled ones by only interacting with NRE models trained on noise labels.
Specifically, for \textit{Gold Label Mix}, we use the same $200$ labels obtained at the pattern-refinement stage as the high-quality labels.
To focus on the impact of training labels produced with different methods,
besides for fixing all hyper-parameters exactly same,
we run the NRE model with five random seeds, ranging from 0 to 4, for each case and present the averaged scores.

\paragraph{Overall Results.}
Table~\ref{table:experiment} shows the overall results with precision (P.), recall (R.) and F1 scores.
For a majority of tasks suffering large labeling noises, including $R_1$, $R_4$, $R_5$, $R_8$, $R_9$ and $R^u_8$, we improve the F1 score by $5.0$ over the best baseline.
Notably, the F1 improvement for task $R_1$ has reached $40$.
For some tasks with fewer noises, including $R_0$, $R_7$, $R^u_7$ and $R^u_9$, our method can obtain small improvements.
For a few tasks, such as $R_3$, $R_6$ and $R^u_6$, only using DS is sufficient to train competitive models.
In such cases, fusing other weak labels may have negative effects, but these side effects are small.
The detailed reasons for these improvements will be elaborated together with the diagnostic results in Section~\ref{sec:pat_diag_res}.
Another interesting observation is that RLRE yields the best result on tasks $R_2$ and $R^u_6$ but gets worse results than the vanilla DS on tasks $R_0$, $R_1$, $R_4$ and $R_7$.
Since the instance selector used in RLRE is difficult to be interpreted, we can hardly figure out the specific reason.
We conjecture that this behavior is due to the gap between maximizing the likelihood of the NRE model and the ground-truth instance selection.
In contrast, DIAG-NRE can contribute both stable and interpretable improvements with the help of human-readable patterns.

\subsection{Pattern-based Diagnostic Results}
\label{sec:pat_diag_res}

Besides for improving the extraction performance, DIAG-NRE can interpret different noise effects caused by DS via refined patterns, as Table~\ref{table:diagnosis} shows.
Next, we elaborate these diagnostic results and the corresponding performance degradation of NRE models from two perspectives: false negatives (FN) and false positives (FP).

\paragraph{FN.}
A typical example of FN is task $R_1$ (\texttt{Administrative\_Division}), where the precision of DS-generated labels is fairly good but the recall is too low.
The underlying reason is that the relational facts stored in the KB cover too few real facts actually contained by the corpus.
This low-recall issue introduces too many negative instances with common relation-supporting patterns and thus confuses the NRE model in capturing correct features.
This issue also explains results of $R_1$ in Table~\ref{table:experiment} that the NRE model trained on DS-generated data achieves high precision but low recall,
while DIAG-NRE with reinforced positive patterns can obtain significant improvements due to much higher recall.
For tasks $R_8$  (\texttt{Birthplace}) and $R_9$  (\texttt{Deathplace}), we observe the similar low-recall issues.

\paragraph{FP.}
The FP errors are mainly caused by the assumption of DS described in the introduction.
For example, the precision of DS-generated labels for tasks $R_8$ and $R^u_8$ is too low.
This low precision means that a large portion of DS-generated positive labels do not indicate the target relation.
Thus, this issue inevitably causes the NRE model to absorb some irrelevant patterns.
This explanation also corresponds to the fact that we have obtained some negative patterns.
By reducing labels with FP errors through negative patterns,
DIAG-NRE can achieve large precision improvements.

For other tasks, DS-generated labels are relatively good,
but the noise issue still exists, major or minor,
except for task $R_3$ (\texttt{Contains}),
where labels automatically generated by DS are incredibly accurate.
We conjecture the reason for such high-quality labeling is that for task $R_3$, the DS assumption is consistent with the written language convention: when mentioning two locations with the containing relation in one sentence, people get used to declaring this relation explicitly.

\begin{figure}[tb]
\centering
\includegraphics[width=0.9\linewidth]{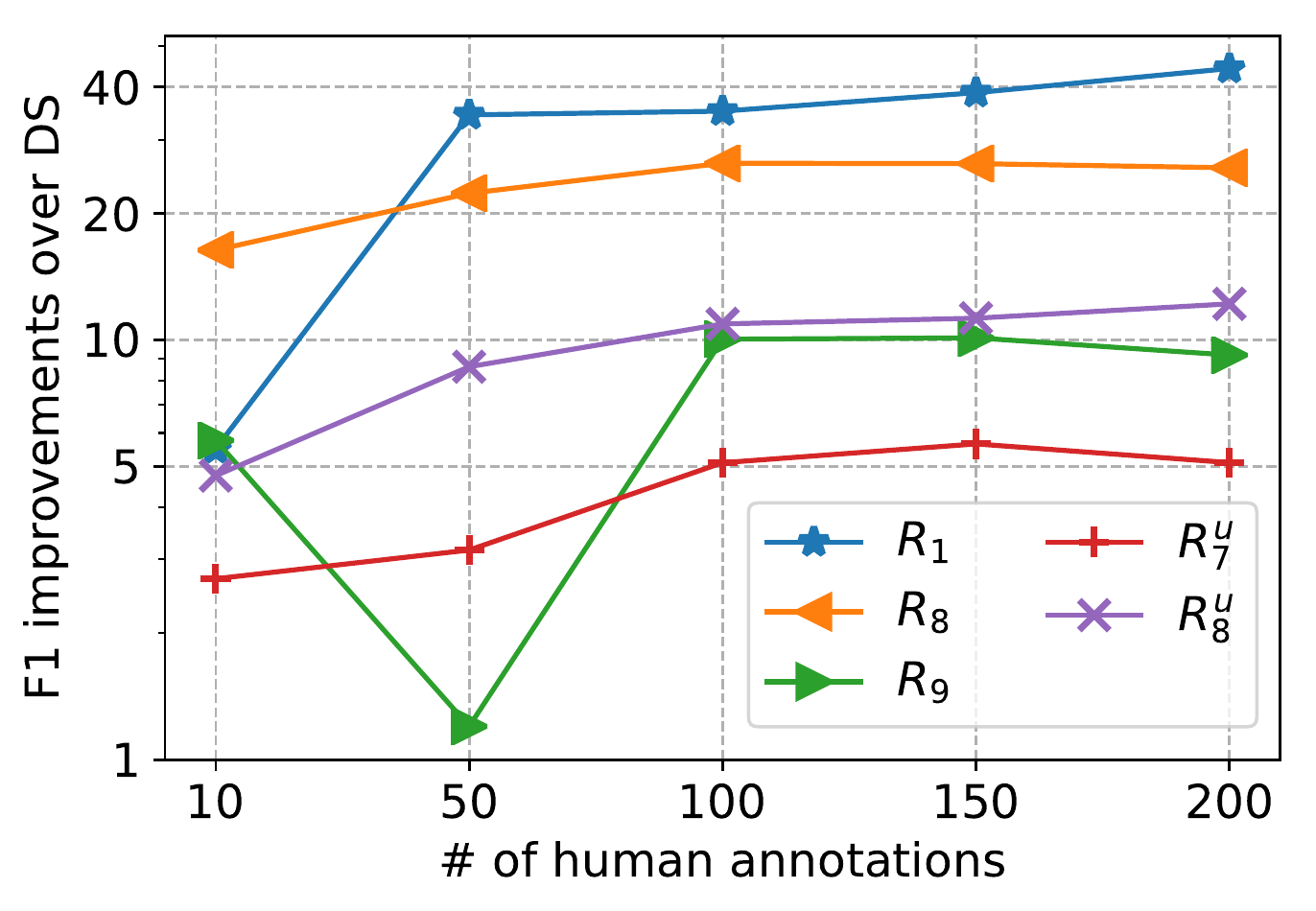}
\caption{The F1 improvements of DIAG-NRE over DS with the increased number of human annotations.}
\label{fig:adap_diag}
\end{figure}

\subsection{Incremental Diagnosis}
In addition to the performance comparisons based on $200$ human-annotated instances, we show the incremental diagnosis ability of DIAG-NRE by gradually increasing the number of human annotations from $10$ to $200$.
As Figure~\ref{fig:adap_diag} shows, where we pick those tasks (three from NYT and two from UW) suffering large labeling noises, most tasks experience a rapid improvement phase with the help of high-quality patterns automatically generated by DIAG-NRE and then enter a saturate phase where adding annotations does not contribute much.
This saturation accords with the intuition that high-quality relational patterns are often limited.
The only exception is task $R_9$ that drops first and then increases again,
the reason is that the fully automatic pattern refinement of DIAG-NRE produces one incorrect pattern accidentally, while later patterns alleviate this mistake.
Actually, in practice, users can further curate patterns generated by DIAG-NRE to get even better results, which can also be much easier and quicker than writing patterns from scratch.

\subsection{Case Studies}
Table~\ref{table:case} shows five pattern examples from three tasks.
For task $R_1$, the positive pattern can remedy the extremely low coverage caused by DS.
For tasks $R_8$ and $R^u_9$, besides for the help of the positive pattern, the negative pattern can correct many FP labels caused by DS.
These cases intuitively illustrate the ability of DIAG-NRE to diagnose and denoise DS-generated labels.


\section{Conclusion and Future Work} 
\label{sec:conclusion}

In this paper, we propose a neural pattern diagnosis framework, DIAG-NRE, to diagnose and improve NRE models trained on DS-generated data.
DIAG-NRE not only eases the hard pattern-writing work of human experts by generating patterns automatically, but also enables the quick generalization to new relation types by only requiring a small number of human annotations.
Coupled with the WLF model, DIAG-NRE can produce denoised labels to retrain a better NRE model.
Extensive experiments with comprehensive analyses demonstrate that DIAG-NRE can contribute both significant and interpretable improvements.

For the future work, we plan to extend DIAG-NRE to other DS-based applications, such as question answering~\cite{lin2018denoising}, event extraction~\cite{chen2017automatically}, etc.

\section*{Acknowledgements}
This work is supported in part by the National Natural Science Foundation of China (NSFC) Grant 61532001, 61572273, 61532010, Tsinghua Initiative Research Program Grant 20151080475, and gift funds from Ant Financial and Nanjing Turing AI Institute.
Xu Han is also supported by 2018 Tencent Rhino-Bird Elite Training Program.

\bibliography{reference-short}
\bibliographystyle{acl_natbib}

\newpage
\appendix

\section{Appendices}
In the appendices, we introduce formulation details of the weak-label-fusion (WLF) model and the hyper-parameters for our neural relation extraction (NRE) model.

\subsection{Weak Label Fusion}
As mentioned in the main body, we employ the \textit{data programming} (DP)~\cite{ratner2016data} as our WLF model.
DP proposed an abstraction of the weak label generator, named as the \textit{labeling function} (LF), which can incorporate both DS and pattern-based heuristics.
Typically, for a binary classification task, an LF is supposed to produce one label (+1: \textit{positive}, -1: \textit{negative} or 0: \textit{unknown}) for each input instance.
In our case, the LF of DS generates +1 or -1, LFs of positive patterns generate +1 or 0, and LFs of negative patterns generate -1 or 0.

Given $m$ labeling functions, we can write the joint probability of weak labels $\bm{L}^s$ and the true label $Y^s \in \{-1,+1\}$ for instance $s$, $P_{\bm{\alpha}, \bm{\beta}}(\bm{L}^s, Y^s)$, as
\begin{align*}
  \frac{1}{2} \prod_{i=1}^m ( \beta_i \alpha_i & \bm{1}_{\{L^s_i = Y^s\}} 
  + \beta_i (1 - \alpha_i) \bm{1}_{\{L^s_i = - Y^s\}} \\
  &+ (1 - \beta_i)  \bm{1}_{\{L^s_i = 0\}} ),
\end{align*}
where each $L^s_i \in \{-1,0,+1\}$ denotes the weak label generated for instance $s$ by the $i^{th}$ labeling function, and $\bm{\alpha}$ and $\bm{\beta}$ are model parameters to be estimated.

Originally, \citet{ratner2016data} conducted the unsupervised parameter estimation based on unlabeled data by solving 
\[
\max_{\bm{\alpha}, \bm{\beta}} \sum_{s \in S} \log \rbr{\sum_{Y^s} P_{\bm{\alpha}, \bm{\beta}}\rbr{\bm{L}^s, Y^s)}}.
\]

Different from the general DP that treats each LF with the equal prior, we have strong priors that patterns produced by DIAG-NRE are either supportive or unsupportive of the target relation with high probabilities.
Therefore, in our case, we directly employ the small labeled set $S_L$ obtained at the pattern-refinement stage to estimate $(\bm{\alpha}, \bm{\beta})$ by solving 
\begin{align*}
    \max_{\bm{\alpha}, \bm{\beta}} 
    \sum_{s \in S_L} \log P_{\bm{\alpha}, \bm{\beta}}(\bm{L}^s, Y^s),
\end{align*}
where the closed-form solutions are
\begin{align*}
    \alpha_i &= \frac{ \sum_{s \in S_L} \bm{1}_{\{L^s_i = Y^s\}} }
    { \sum_{s \in S_L} \sbr{ \bm{1}_{\{L^s_i = Y^s\}} + \bm{1}_{\{L^s_i = -Y^s\}} } }, \\
    \beta_i &= \frac{ \sum_{s \in S_L} \sbr{ \bm{1}_{\{L^s_i = Y^s\}} + \bm{1}_{\{L^s_i = -Y^s\}} } }{ |S_L| }, 
\end{align*}
for each $i \in \{1, \cdots, m\}$.
After estimating these parameters, we can infer the true label distribution by the posterior $P_{\bm{\alpha}, \bm{\beta}}(Y^s | \bm{L}^s)$ and use the denoised soft label to train a better NRE model, just as~\citet{ratner2016data} did.

\subsection{Hyper-parameters of the NRE model}
For the NRE model, we implement a simple yet effective LSTM-based architecture described in~\cite{zhou2016attention}.
We conduct the hyper-parameter search via cross-validation and adopt the following configurations that can produce pretty good results for all $14$ tasks.
First, the word embedding table ($d^w = 100$) is initialized with Glove vectors~\cite{pennington2014glove},
the size of the position vector ($d^p$) is 5,
the maximum length of the encoded relative distance is $60$, 
and we follow~\cite{zeng2015distant,lin2016neural} to randomly initialize these position vectors. 
Besides, the LSTM hidden size is 200, and the dropout probabilities at the embedding layer, the LSTM layer and the last layer are 0.3, 0.3 and 0.5, respectively.
During training, we employ the Adam~\cite{kingma2014adam} optimizer with the learning rate of 0.001 and the batch size of 50.
Moreover, we select the best epoch according to the score on the validation set.

Notably, we observe that when training on data with large labeling noises, different parameter initializations can heavily influence the extraction performance of trained models.
Therefore, as mentioned in the main body, to clearly and fairly show the actual impact of different types of training labels, we restart the training of NRE models with $5$ random seeds, ranging from $0$ to $4$, for each case and report the averaged scores.

\end{document}